\documentclass[conference]{IEEEtran}
\IEEEoverridecommandlockouts
\usepackage{cite}
\usepackage{amsmath,amssymb,amsfonts}
\usepackage{algorithmic}
\usepackage{graphicx}
\usepackage{textcomp}
\usepackage{xcolor}

\newtheorem{definition}{Definition}
\usepackage[para,online,flushleft]{threeparttable}
\usepackage{tikz}
\usepackage{subcaption}
\usepackage{multirow}
\usepackage{hyperref}

\def\BibTeX{{\rm B\kern-.05em{\sc i\kern-.025em b}\kern-.08em
    T\kern-.1667em\lower.7ex\hbox{E}\kern-.125emX}}

\begin{document}

\title{Inductive Learning on Commonsense Knowledge Graph Completion}

\author{
\IEEEauthorblockN{Bin Wang}
\IEEEauthorblockA{\textit{Electrical and Computer Engineering} \\
\textit{University of Southern California}\\
Los Angeles, CA, USA \\
bwang28c@gmail.com}
\and
\IEEEauthorblockN{Guangtao Wang}
\IEEEauthorblockA{\textit{\hspace{2cm}AI Research\hspace{2cm}} \\
\textit{JD.com}\\
Mountain View, CA, USA \\
guangtao.wang@jd.com}
\and
\IEEEauthorblockN{Jing Huang}
\IEEEauthorblockA{\textit{\hspace{2cm}AI Research\hspace{2cm}} \\
\textit{JD.com}\\
Mountain View, CA, USA \\
jing.huang@jd.com}
\and
\IEEEauthorblockN{Jiaxuan You}
\IEEEauthorblockA{\textit{\hspace{1.7cm}Computer Science\hspace{1.7cm}} \\
\textit{Stanford University}\\
Stanford, CA, USA \\
jiaxuan@stanford.edu}
\and
\IEEEauthorblockN{Jure Leskovec}
\IEEEauthorblockA{\textit{\hspace{1.6cm}Computer Science\hspace{1.6cm}} \\
\textit{Stanford University}\\
Stanford, CA, USA \\
jure@stanford.edu}
\and
\IEEEauthorblockN{C.-C. Jay Kuo}
\IEEEauthorblockA{\textit{Electrical and Computer Engineering} \\
\textit{University of Southern California}\\
Los Angeles, CA, USA \\
cckuo@sipi.usc.edu}
}

\maketitle

\begin{abstract}
Commonsense knowledge graph (CKG) is a special type of knowledge graph (KG), where entities are composed of free-form text. Existing CKG completion methods focus on transductive learning setting, where all the entities are present during training. Here, we propose the first inductive learning setting for CKG completion, where unseen entities may appear at test time. We emphasize that the inductive learning setting is crucial for CKGs, because unseen entities are frequently introduced due to the fact that CKGs are dynamic and highly sparse. We propose InductivE as the first framework targeted at the inductive CKG completion task. InductivE first ensures the inductive learning capability by directly computing entity embeddings from raw entity attributes. Second, a graph neural network with novel densification process is proposed to further enhance unseen entity representation with neighboring structural information. 
Experimental results show that InductivE performs especially well on inductive scenarios where it achieves above 48\% improvement over previous methods while also outperforms state-of-the-art baselines in transductive settings.
\end{abstract}


\section{Introduction}
Knowledge graphs (KGs) are represented as triplets where entities
(nodes) are connected by relationships (edges). It is a structured
knowledge base with various applications such as recommendation
systems, question answering, and natural language understanding
\cite{wang2018dkn,cui2019kbqa,liu2020k}. In practice, most KGs are far from complete. Therefore, 
predicting missing facts is one of the most fundamental problems in this
field. A lot of embedding-based methods have been proposed and shown to
be effective on the KG completion task \cite{bordes2013translating,trouillon2016complex,convE,sun2019rotate,balazevic-etal-2019-tucker,Bansal2019A2NAT,tang2019orthogonal}. However, relatively little work targets at
commonsense knowledge graph (CKG) completion. There are unique
challenges encountered by applying existing KG embedding methods to
CKGs (e.g. ConceptNet \cite{speer2013conceptnet} and ATOMIC
\cite{sap2019atomic}). 
    
\begin{figure}[tb]
\centering
\includegraphics[width=0.5\textwidth]{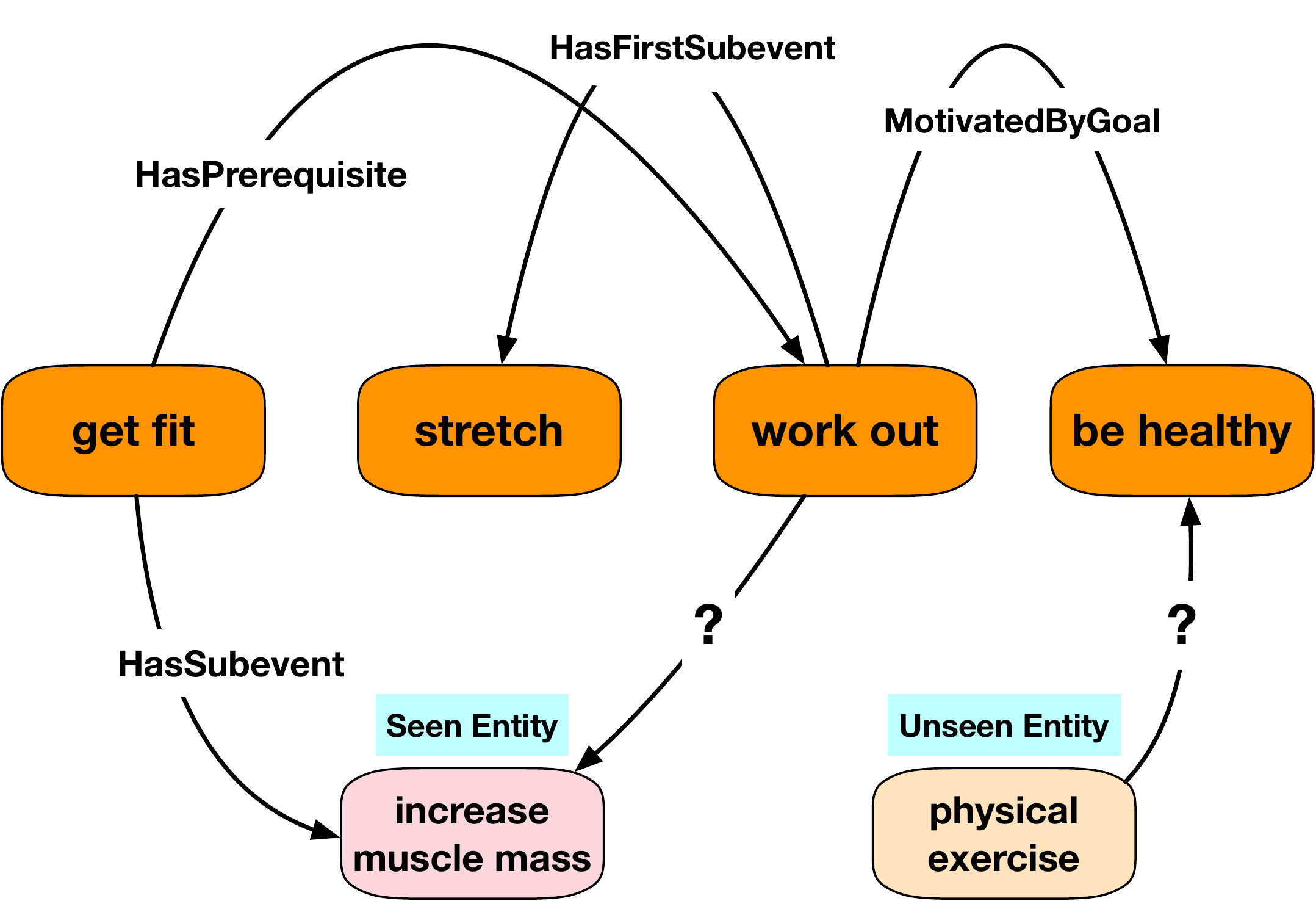}
\caption{Illustration of a fragment of ConceptNet commonsense knowledge graph: Link prediction for seen entities (transductive) and unseen entities (inductive).
} \label{fig:cn_example}
\end{figure}

First, many real-world CKGs are dynamic in nature, and entities with unseen text/names are introduced from time to time. We call these entities unseen entities because they are not involved in training and only appear in testing.

Second, entity attributes of CKGs are composed of free-form texts which are not present in non-attributed KG datasets. As shown in Figure \ref{fig:cn_example}, entity description has rich semantic meaning and commonsense knowledge can be largely inferred from their implicit semantic relations. However, we notice that 
often entities refer to the same concept are stored as distinct ones, resulting in the graph to be larger and sparser. 
As shown in Table 1, the average in-degree of ConceptNet and ATOMIC is only $1/15$ and $1/8$ comparing with that of FB15K-237, a popular KG dataset.
Since CKGs are highly sparse and can be disconnected, a portion of entities are isolated from the main graph structure. These entities are also unseen entities and how to obtain embeddings for these isolated entities remains challenging. 

Therefore, the inductive learning problem on commonsense knowledge graph completion is particularly important with practical necessities. An example for CKG completion is shown in Figure \ref{fig:cn_example}. Transductive setting targets on predicting missing links for seen entities. The predictions can be made from two perspectives: 1) entity attributes and 2) existing links for seen entities. In contrast, inductive setting works on purely unseen entities where only entity attributes can be leveraged in the first place. Different from previous inductive setting on non-attributed KGs \cite{teru2019inductive,albooyeh2020out}, inductive learning on CKG assumes unseen entities are purely isolated and does not have any existing links. Therefore, this problem is unique and remains unexplored.


Many existing KG embedding models are focusing on non-attributed graphs with entity embeddings obtained during training \cite{bordes2013translating,sun2019rotate,tang2019orthogonal} or using attribute embedding solely for initialization \cite{malaviya2020commonsense}. Therefore, all entities are required to present in training. Otherwise, the system will not have their embeddings and hence transductive as originally proposed.





In this work, we first propose and define the inductive learning problem on CKG completion. Then, an inductive learning framework called InductivE is introduced to address the above-mentioned challenges and several components are specially designed to enhance its inductive learning capabilities. First, the inductive capability of InductivE is guaranteed by directly building representations from entity descriptions, not merely using entity textual representation as training initialization. Second, a novel graph encoder with densification is proposed to leverage structural information for unseen entities and entities with limited neighboring connections. Overall, InductivE follows an encoder-decoder framework to learn from both semantic representations and updated graph structures.





The main contributions can be summarized as follows. \footnote{Our code is now publicly available:  \href{https://github.com/BinWang28/InductivE}{here}.}
\begin{itemize}
\item[1.] A formal definition of inductive learning on CKGs is
presented. We propose the first benchmark for inductive CKG completion task, including new data splits and testing schema, to facilitate future research.
\item[2.]
InductivE is the first model
that is dedicated to inductive learning on commonsense knowledge graphs.
It leverages entity attributes based on transfer learning from word
embedding and graph structures based on an novel graph neural network with densification.
\item[3.] Comprehensive experiments are conducted on ConceptNet and ATOMIC datasets. The improvements are demonstrated in both transductive and inductive settings. InductivE performs especially well on inductive learning scenarios with an over 48\% improvement on MRR comparing with previous methods.
\end{itemize}

\section{Problem Definition}

\begin{definition} \textbf{Commonsense Knowledge Graph} (CKG) is represented by $G=(V,E,R)$, where $V$ is the set of nodes/entities, $E$ is the set of edges and $R$ is the set of relations. Edges consist of triplets $(h,r,t)$ where head entity $h$ and tail entity $t$ are connected by relation $r$: $E=\{(h,r,t)|h\in V, t\in V, r\in R\}$, and each node comes with a free-text description.
\end{definition}
    

\begin{definition}\textbf{CKG Completion}
Given a commonsense knowledge graph $G=(V,E,R)$, CKG completion is defined as the task of predicting missing triplets $E'=\{(h,r,t)|(h,r,t)\not\in E\}$. It includes both transductive and inductive settings. Transductive CKG completion is defined as predicting missing triplets $E''=\{(h,r,t)|(h,r,t)\not\in E,$ $h \in V, t \in V, r \in R\}$. 
\end{definition}

\begin{definition} \textbf{Inductive CKG completion} is defined as predicting missing triplets $E'''=\{(h,r,t)|(h,r,t)\not\in E,$ $h\in V'\ or\ t\in V', r\in R\}$, where $V' \cap V = \emptyset$ and $V' \neq \emptyset$.
\end{definition}

        \begin{table*}[htb]
        \centering
        \caption{Statistics of CKG datasets. Unseen Entity \% indicates the percentages of 
        unseen entities in all test entities. 
        }
        \resizebox{\textwidth}{!}{
        \begin{threeparttable}
        \begin{tabular}{ c | c | c | c | c | c | c | c}
        \hline
         Dataset        & Entities  & Relations & Train Edges   & Valid Edges   & Test Edges   & Avg. In-Degree    & Unseen Entity \%    \\ \hline
         CN-100K        & 78,334     & 34        & 100,000        & 1,200          & 1,200         & 1.31              & 6.7\%             \\ \hline
         CN-82K         & 78,334     & 34        & 81,920         & 10,240         & 10,240        & 1.31              & 52.3\%            \\ \hline
         ATOMIC         & 304,388    & 9         & 610,536        & 87,700         & 87,701        & 2.58              & 37.6\%            \\ \hline 
        
        \end{tabular}
        \begin{tablenotes}
        \item[$\star$] For comparison, a popular KG dataset: FB-15K-237 has 14,541 entities, 310,115 edges and 18.76 Avg. In-Degree. 
        \item[$\star$] In transductive settings, all entities are assumed visible during training including unseen entities. In contrast, unseen entities can only be used during testing stage for inductive settings.
        \end{tablenotes}
        \end{threeparttable}
        }
        \label{tab:statistics}
        \end{table*}

        \begin{figure}[htb]
            \centering
            \begin{subfigure}[t]{0.25\textwidth}
                \centering
                \includegraphics[width=\textwidth]{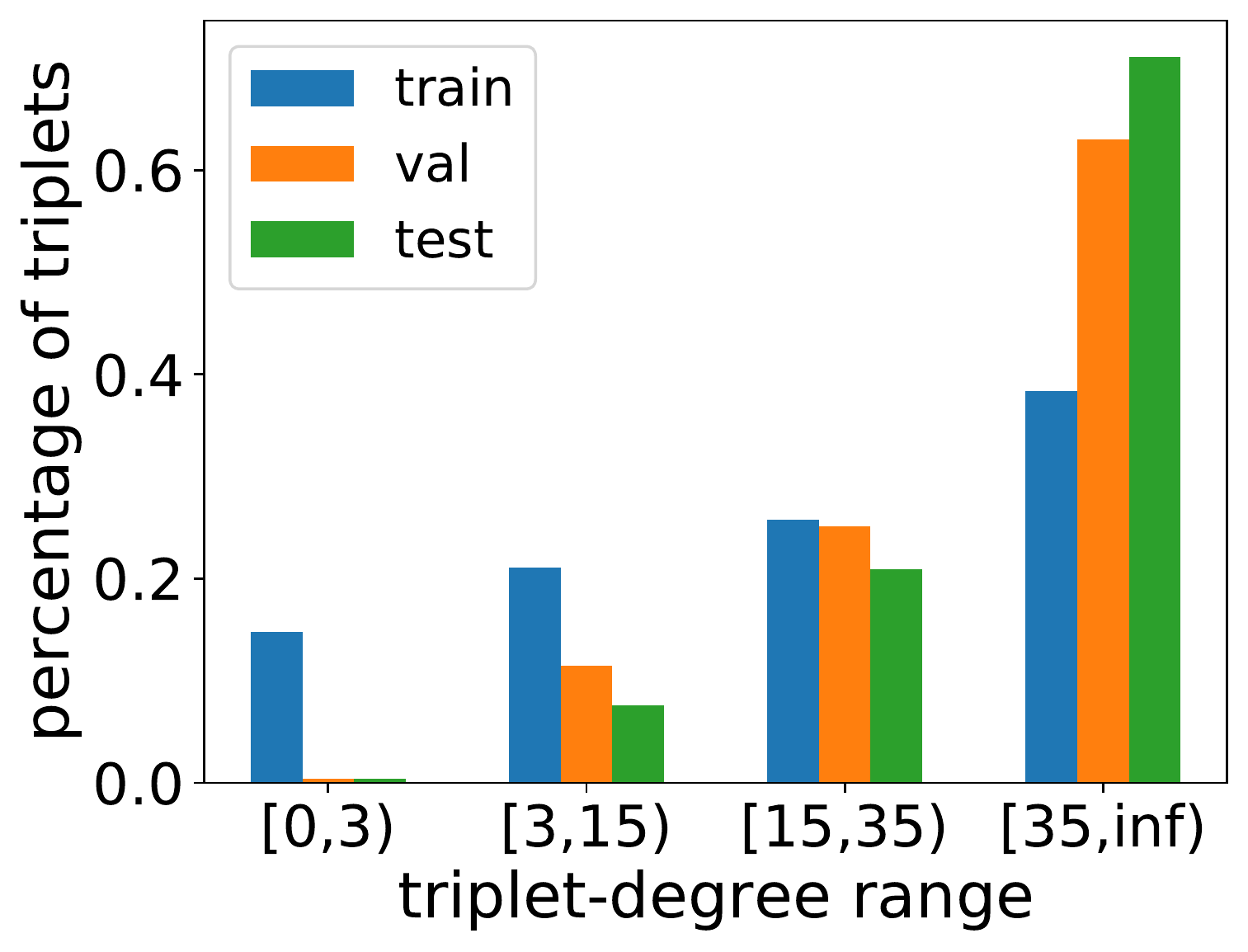}
                \caption{CN-100K}
            \end{subfigure}%
            \begin{subfigure}[t]{0.25\textwidth}
                \centering
                \includegraphics[width=\textwidth]{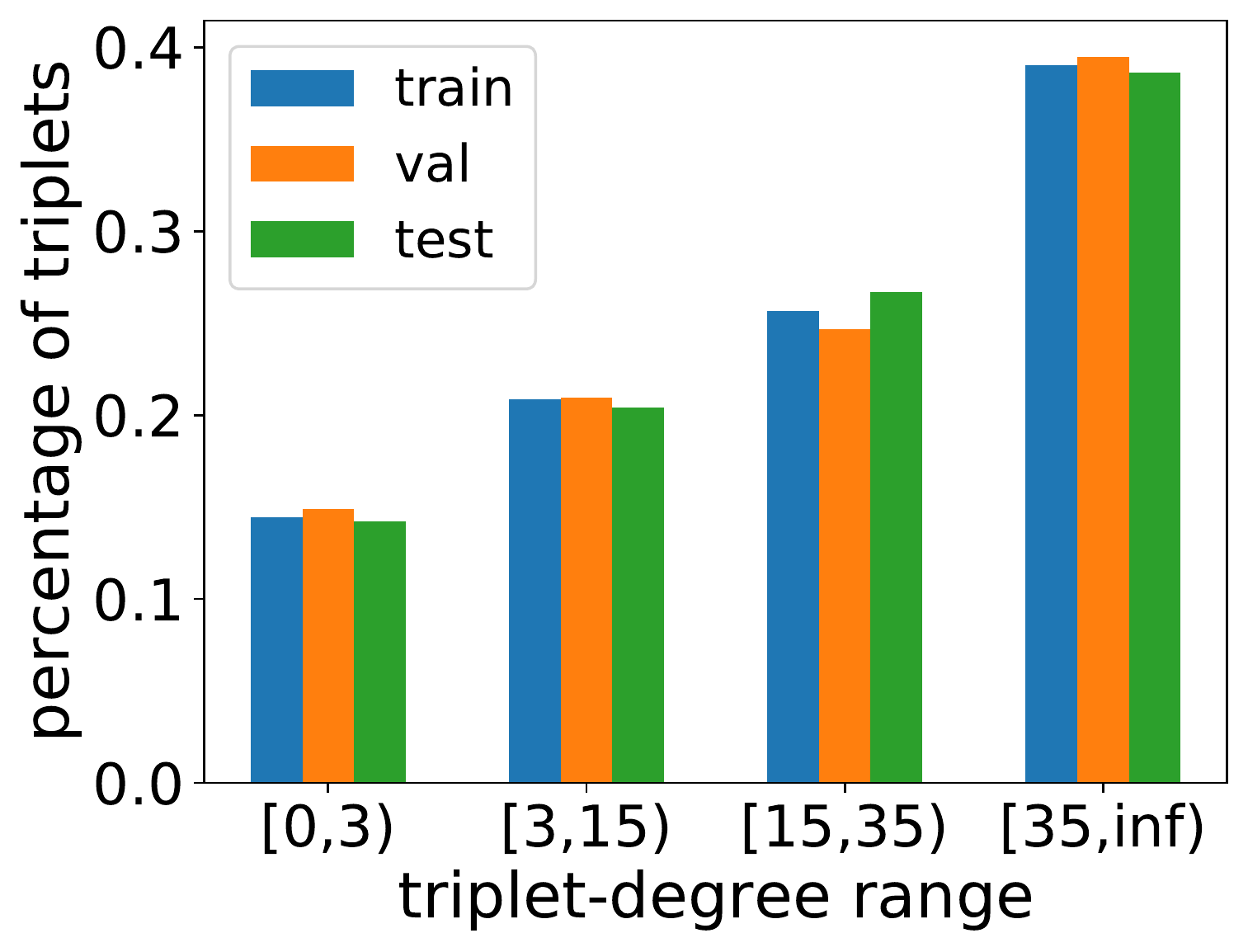}
                \caption{CN-82K}
            \end{subfigure}
            \caption{The triplet-degree distribution for the ConceptNet dataset, where the triplet-degree is the average of head and tail degrees. Triplets with high degrees are easier to predict in general. CN-100K split is clearly 
            unbalanced comparing with CN-82K.} \label{fig:data}
        \end{figure}

\section{Dataset Preparation}

    Three CKG datasets are used to evaluate the link prediction (i.e., CKG completion) task.
    Their
    statistics are shown in Table \ref{tab:statistics}. CN-82K is our newly proposed split for better evaluation on link prediction task.
    Besides the standard split, we create an inductive split for CN-82K and ATOMIC called CN-82K-Ind and ATOMIC-Ind to specifically evaluate model's generalizability for unseen entities.

\subsection{Standard Split: CN-100K, CN-82K, ATOMIC}
        
            CN-100K was first introduced by
            \cite{li2016commonsense}.  It contains Open Mind Common Sense (OMCS)
            entries in the ConceptNet 5 dataset \cite{speer2013conceptnet}. ``100K''
            indicates the number of samples in the training data. In the ConceptNet 5
            dataset, each triplet is associated with a confidence score that
            indicates the degree of trust. In the original split of CN-100K, the
            most confident 1,200 triplets are selected for testing and the next
            1,200 most confident triplets are used for validation. Entities have a
            text description with an average of 2.9 words. 
        
            CN-100K was originally
            proposed to separate true and false triplets.
            It is not ideal for link prediction for two reasons.
            First, its data split ratio is biased. For 100,000 training samples,
            CN-100K contains only 2,400 triplets (2.4\%) for validation (1.2\%) and testing (1.2\%).
            With such limited testing and validation sets, we
            see a large variance in evaluation. Second, the testing and the
            validation sets are selected as the most confident samples which are the least challenging ones.
            As shown in Figure \ref{fig:data}(a), we see an
            unbalanced distribution in triplet-degree among train, validation and test
            sets. In order to test the performance of the link prediction task, we
            create and release a new data split called CN-82K. 
            
            CN-82K is a uniformly sampled
            version of the CN-100K dataset. 80\% of the triplets are used for the
            training, 10\% for validation, and the remaining 10\% for testing. As shown
            in Figure \ref{fig:data}(b), CN-82K is more balanced among train,
            validation and test sets w.r.t the triplet-degree distribution.  
            
            ATOMIC contains everyday
            commonsense knowledge entities organized as \textit{if-then} relations
            \cite{sap2019atomic}.  It contains over 300K entities in total and
            entities are composed of text descriptions with an average of 4.4 words.
            Here, we use the same data split as done in
            \cite{malaviya2020commonsense}. Different from the ATOMIC split in \cite{Bosselut2019COMETCT,sap2019atomic}, the data split is a random split with ratio 80\%/10\%/10\% for train/valid/test. Therefore, there is still a reasonable percentage of entities are unseen during training.
            
            
        \subsection{Inductive Split: CN-82K-Ind, ATOMIC-Ind}
        
            To evaluate a model's generalizability to unseen entities, we create new validation and test sets for CN-82K and ATOMIC, while keeping the training set unchanged. The inductive validation and testing sets are subsets of the original validation and test sets, which contain only triplets with at least one unseen entities. After filtering, for CN-82K-Ind,
            the validation and test sets contain 5,715 and 5,655 triplets,
            respectively. For ATOMIC-Ind, the validation and test sets contain 24,355 and 24,486
            triplets, respectively. We show by experiments that previous models fail in such scenarios (please refer to Table~\ref{tab:inductive}) and propose a new benchmark called InductivE for inductive CKG completion.
            

\begin{figure*}[tb]
\centering
\includegraphics[width=\textwidth]{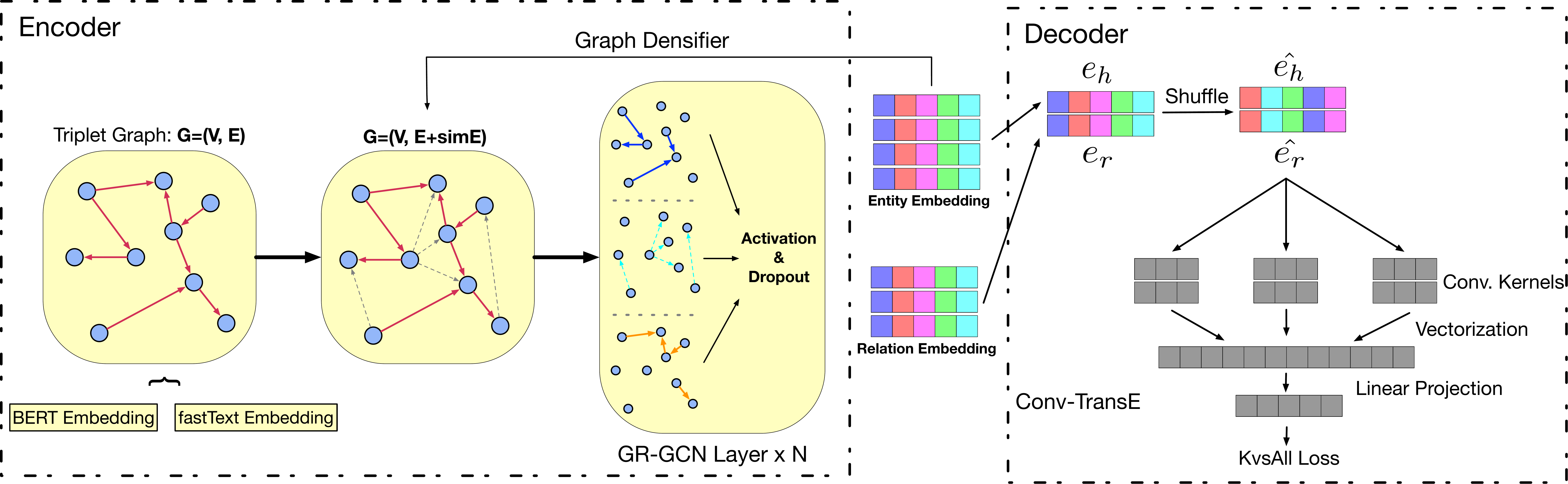}
\caption{Illustration of the proposed InductivE model. The encoder contains
multiple GR-GCN layers applied to the graph to aggregate node's local
information. The decoder uses an enhanced Conv-TransE model.}\label{fig:m1}
\end{figure*}

\section{InductivE}

In this section, we introduce InductivE, the first benchmark of inductive learning on CKGs, which includes a free-text encoder, a graph encoder with densification and a convolutional decoder. The overall architecture is illustrated in Figure \ref{fig:m1}.

\subsection{Free-Text Encoder}

Our free-text encoder embeds text attributes with a pre-trained language model and word-embedding. For pre-trained language model, BERT \cite{devlin-etal-2019-bert} is applied and further fine-tuned on entity textual attributes, which allows BERT to better model our domain-specific data. 
For word embedding, fastText model is used and mean-pooling is applied when the text sequence contains more than one word \cite{bojanowski2017enriching,wang2019evaluating}. Finally, two representations are concatenated together as the final entity representation.
For inductive purpose, the free-text encoder is viewed as a feature extractor and entity embeddings are fixed during training. 

BERT shows superior performance in many NLP tasks. However, for CKG datasets, fastText embedding exhibits comparable performance. We infer that fastText can be close to BERT feature when handling short text sequences. More discussion is provided in Sec. \ref{sec:abla}.

\subsection{Graph Encoder}

    In addition to semantic representation, it is also desired to investigate the possibility to enhance inductive ability with entity neighboring structures. However, it is challenging to leverage structural information for unseen entities because, as isolated ones, unseen entities do not contain any existing links at the beginning. Therefore, we first densify the graph by add similarity links using a novel graph densifier. Then, a gated-relational graph convolutional network is applied to learn from the graph structural information.
    
    

    \subsubsection{Graph densifier}
        
        We observe that some entities often share the same ontological concept in
        CKGs. For example, \emph{``work out''} and \emph{``physical exercise''}
        in Figure  \ref{fig:cn_example} are similar concepts but presented as two
        distinct entities. Similarity edges are added to densify
        the graph to provide more interactions between similar concepts. 

        
        Here, we propose a novel \emph{graph densifier} to generate high-quality links among semantic-similar entities.
        We first compute the pairwise cosine similarity using the output entity feature of
        our graph encoder rather than original entity features. Then, for each node $i$, we identify $k_i$ nearest neighbors and
        add directed similarity edges from them to node $i$ to densify the graph.
        To balance the resulting node degree across the graph, the
        number of added synthetic edges is node-dependent. More synthetic edges will be added to nodes with fewer connections. Specifically, the number of added edges $k_i$ for each node $i$
        is determined by
        \begin{equation} \label{eqn7}
        k_i = 
        \begin{cases}
         0,             & \text{if $m \leq degree(i)$}, \\
         m-degree(i)    & \text{otherwise},
        \end{cases}  
        \end{equation}
        where $m$ is a hyperparameter to determine the number of similarity
        edges added for each node, and $degree(i)$ represents the degree of node $i$.  The densified graph is updated periodically during
        training based on the above-mentioned scheme.  
        
        
        
        In the testing stage, to infer the embedding for unseen entities, we first get entity representations by applying our trained graph encoder on the CKG graph, in which the unseen entities are isolated; then similarity edges are added according to these representations to densify the graph; finally, the densified graph is encoded again using our trained graph encoder to get final entity embeddings, which serve as the input to the decoder.
        
        The logic behind our densification design is as follows: first, we did not use raw feature with threshold because a lower threshold leads to noisy edges and a higher threshold provides little extra information as the feature already serves as the input; second, more edges are added to unseen entities to ensure enough structural information are incorporated for inductive purposes.

    \subsubsection{Gated-Relational GCN}

        R-GCN \cite{schlichtkrull2018modeling} is effective in learning the node representation over graph with relational information between nodes (e.g., Knowledge graph).
        Our model is an extension of the R-GCN \cite{schlichtkrull2018modeling}
        model.
        First, in CKGs, the neighboring conditions can vary a lot from node to node. It is desired to adaptively control the amount of information fused to the center node from their neighboring connections. Therefore, a gating function is added to R-GCN for this purpose based on the interaction of the center and neighboring nodes.
        Second, to increase the efficiency of R-GCN model, instead of using relation-type-specific transformation matrices, one unified transformation matrix $W_1$ is adopted for all neighboring relation types. 
        As a result, our graph encoder is composed of multiple newly proposed gated-relational graph convolutional (GR-GCN) layers.
        
        The first convolutional layer takes the output from free-text encoder as the input: $h_i^{(0)}=x_i$.
        At each layer, the update message is a weighted sum of
        a transformation of center node $u_c$ and a transformation of its
        neighbors $u_n$ in form of
        \begin{equation} \label{eqn1}
        h_i^{(l+1)} = \sigma (u_c^{(l)} \odot \beta_i^{(l)} + u_n^{(l)} \odot
        (1-\beta_i^{(l)})),
        \end{equation}
        where $\beta_i$ denotes a gating function, $\odot$ denotes an
        element-wise multiplication, $\sigma(\cdot)$ is a nonlinear activation
        function.
        The two transformations in Eq. (\ref{eqn1}) are defined as
        \begin{eqnarray} 
        u_c^{(l)} & = & W_o^{(l)} h_i^{(l)}, \label{eqn2} \\
        u_n^{(l)} & = & \sum_{r\in R} \sum_{j\in N_i^r} \frac{1}{|N_i|} 
        \alpha_r^{(l)} W_1^{(l)} h_j^{(l)},  \label{eqn3}
        \end{eqnarray}
        where $N_i^r$ denotes neighbors of node $i$ with relation type $r$, $R$
        denotes all relation types, $\alpha_r^{(l)}$ is the relation weight at
        layer $l$. For all neighboring nodes, we use one unified transformation
        $W_1^{(l)}$, which differs from the transformation of the self-loop
        message denoted by $W_o^{(l)}$, to account for the relation gap between
        neighborhood information and self-connection. 
                
        A gating mechanism controls the amount of the neighborhood message
        flowing into the center node. In this work, a learnable gating
        function is used, which takes both $u_c^{(l)}$ and $u_n^{(l)}$ as the input. It
        can be written as
        \begin{equation} \label{eqn4}
        \beta_i^{(l)} = \mbox{sigmoid} (f([u_c^{(l)},u_n^{(l)}])),
        \end{equation}
        where $[,]$ is the concatenation of self-loop and neighboring messages
        and $f$ is a linear transformation. Finally, a sigmoid function is
        used to ensure $0<\beta_i^{(l)}<1$.

\subsection{Decoder - Conv-TransE}
        
Convolutional decoder is effective in scoring triplets in KGs with high parameter efficiency \cite{convE}.  
Conv-TransE
\cite{shang2019end} is a simplified version of ConvE \cite{convE}.  It
removes the reshaping process before the convolution operation and use 1D convolution operation instead of 2D. Yet, the 2D convolution increases the expressive power of the ConvE model since it allows more interactions between
embeddings as discussed in \cite{convE}. InteractE was proposed
recently to allow even more interactions \cite{vashishth2020interacte}.


As shown in Figure \ref{fig:m1}, we deploy an improved Conv-TransE model that adds a shuffling operation before convolution to enable more interactions across dimensions inspired by InteractE~\cite{vashishth2020interacte}. 
Formally, let $\phi_s$
represent the horizontal shuffling process, the score function in
our decoder is defined as:
\begin{equation} \label{eqn5}
\begin{split}
score&(e_h,e_r,e_t)  =  \\
& f(vec(f(\phi_s([e_h;e_r])*w))W))e_t 
\end{split}
\end{equation}
where $e_h/e_t \in \mathbb{R}^d$ is the head/tail embedding and $e_r \in
\mathbb{R}^d$ is the relation embedding. 
These entity embeddings come from the
output of the graph encoder. The relation embedding is randomly
initialized and jointly trained with the end-to-end framework. 
$f$ denotes a non-linear activation function.
In a feed-forward pass, $e_h$ and $e_r$ are first stacked as a $\{2\times d\}$ matrix and shuffled horizontally. It is used as the input to a 1D convolutional layer with filters $w$. Each filter is with 1D convolutional kernel with size $\{2\times n\}$. The output is further reshaped as a vector and projected back to the original $d$ dimensions using a linear transformation. Finally, an inner product with the tail embedding $e_t$ is performed in the $d$-dimensional space as the final triplet score. 


To train our model parameter, we use KvsAll
training schema \cite{convE} by considering all entities simultaneously. Instead of scoring each triplet $(h,r,t)$, we take each $(h,r)$ pair and score it with all entities (positive or negative) as tail. Some pairs could have more than one positive tail entities. Thus, it is a multi-label problem, for which we adopt the binary cross-entropy loss:
\begin{equation} \label{eqn6}
L(p,t) = - \frac{1}{N} \sum_i (t_i\cdot log(p_i)+(1-t_i)\cdot log(1-p_i)),
\end{equation}
where $t$ is the true label vector with dimension $\mathbb{R}^N$. We get the probability of ($h, r, t$) being positive as $p_i=\mbox{sigmoid}(score(e_h,e_r,e_i))$.

\section{Experiments}

        \begin{table*}[htb]
        \centering
        \caption{Comparison of CKG completion results on CN-100K, CN-82K and ATOMIC datasets. Improvement is computed by comparing with \cite{malaviya2020commonsense}.
        }
        \resizebox{1.0\textwidth}{!}{
        \begin{tabular}{ c | c  c c | c c c | c c c }\hline
             \multirow{2}{*}{\textbf{Model}}                       & \multicolumn{3}{c |}{\textbf{CN-100K}}             & \multicolumn{3}{c |}{\textbf{CN-82K}} & \multicolumn{3}{c}{\textbf{ATOMIC}}  \\
                     & MRR    & Hits@3     & Hits@10   & MRR     & Hits@3             & Hits@10 & MRR        & Hits@3        & Hits@10         \\ \hline
         DistMult   & 10.62  & 10.94 & 22.54    & 2.80  & 2.90   & 5.60   & 12.39  & \textbf{15.18}   & 18.30    \\ 
         ComplEx    & 11.52   & 12.40 & 20.31  & 2.60  & 2.70   & 5.00 & \textbf{14.24} & 14.13  & 15.96     \\ 
         ConvE   & 20.88   & 22.91  & 34.02  & 8.01  & 8.67  & 13.13 & 10.07  & 10.29  & 13.37        \\ 
         RotatE    & 24.72  & 28.20 & 45.41 & 5.71   & 6.00 & 11.02 & 11.16 & 11.54  & 15.60        \\ \hline
         COMET        & 6.07    & 2.92     & 21.17     & -       & -             & -  & 4.91   & 2.40  & \textbf{21.60}   \\ 
         Malaviya et al.    & 52.25  & 58.46  & 73.50     & 16.26 & 17.95  & 27.51     & 13.88 & 14.44   & 18.38    \\ \hline
         \textbf{InductivE}   & \textbf{57.35}  & \textbf{64.50}        & \textbf{78.00}     & \textbf{20.35}   & \textbf{22.65}           & \textbf{33.86}     & 14.21 & 14.82   & 20.57  \\ 
         Improvement            & 9.8\%   & 10.3\%    & 6.1\%     & 25.2\%   & 26.2\%   & 23.1\%    & 2.38\% & 2.63\%  & 11.92\%       \\ \hline
         \end{tabular}
         }
        \label{tab:transductive}
        \end{table*}
        
        \begin{table}[htb]
        \centering
        \caption{Comparison of CKG completion results on unseen entities for
        CN-82K-Ind and ATOMIC-Ind.}
        \resizebox{0.5\textwidth}{!}{
        \begin{tabular}{ c | c   c | c c  }\hline
                             \multirow{2}{*}{\textbf{Model}}        & \multicolumn{2}{c |}{\textbf{CN-82K-Ind}}                  & \multicolumn{2}{c}{\textbf{ATOMIC-Ind}}  \\
                                                            & MRR  & Hits@10   & MRR           & Hits@10     \\ \hline
        ConvE & 0.21  & 0.40 & 0.08  & 0.09 \\
         RotatE &  0.32  & 0.50 & 0.10 & 0.12 \\
        Malaviya et al. & 12.29 & 19.36 & 0.02  & 0.07 \\ \hline
         \textbf{InductivE}     &  \textbf{18.15}  & \textbf{29.37}  &\textbf{2.51} & \textbf{5.45}  \\ \hline
        \end{tabular}
        }
        \label{tab:inductive}
        \end{table}

    \subsection{Experimental Setup}
        
            \subsubsection{Evaluation protocol}
            We use link prediction task with standard evaluation metrics including Mean Reciprocal Rank (MRR) and Hits@10, to evaluate CKG completion models. 
            We report the average results in percentage
            with five runs. Following \cite{convE,sun2019rotate}, each triplet $(h,r,t)$ is measured in two directions: $(h,r,?)$ and $(t,r^{-1},?)$. Inverse relations $r^{-1}$ are added as new relation types and the filtered
            setting is used to filter out all valid triplets before ranking.
        
        \subsubsection{Hyper-parameter settings} Our graph encoder consists of two GR-GCN layers with hidden dimension $500$. The hyperparameter $m$ used in graph densifier is set to $5$ for ConceptNet and $3$ for ATOMIC. The graph is updated every 100 epochs for ConceptNet and 500 epochs for ATOMIC. For convolutional decoder, we use $300$ kernels of size \{$2\times5$\}. For model training, we set the initial learning rate to 3e-4 and 1e-4 for ConceptNet and ATOMIC, respectively, and halve the learning rate when validation metric stops increasing for three times.
        The checkpoint with the highest Mean Reciprocal Rank (MRR) on validation set is used for testing.\footnote{More details are presented in the our released project page.}
        
        Training GCN on large graphs 
        demands large memory space to keep the entire graph parameters.
        Here, for efficiency purposes, we perform uniform random sampling on nodes at each training epoch with sampling size $50$k in all experiments.

        \subsubsection{Baselines}
            We compare with several
            representative models, including DistMult \cite{yang2014embedding},
            ComplEx \cite{trouillon2016complex}, RotatE \cite{sun2019rotate}, ConvE
            \cite{convE}, Malaviya \cite{malaviya2020commonsense} and COMET
            \cite{Bosselut2019COMETCT}. The first four models are competitive KG embedding models without using entity textual attributes. The last two models are focusing on CKGs and also take entity textual attributes into consideration. For non-inductive models, we allow the presence of unseen entities during training in order to perform evaluation. We use their respective official implementations to obtain the baseline results and tune several hyperparameters including batch size, learning rate and embedding dimensions. Finally, our obtained baseline results are compared with the best results reported in existing literature and a higher one is reported.


            
            
            
    \subsection{Result and Analysis} 
    
        \subsubsection{Transductive link prediction}
        
            Table
            \ref{tab:transductive} summarizes results on CN-100K, CN-82K and ATOMIC.
            For
            CN-100K, our model outperforms previous state-of-the-art by 9.8\% on
            MRR and 6.1\% on HITs@10. In
            contrast, the performance for CN-82K is much lower since CN-82K has more
            unseen entities in the testing as shown in Table \ref{tab:statistics}.
            Over 50\% of all entities in testing are unseen. This is very
            challenging for all existing methods. InductivE can learn high-quality
            embedding for all entities and outperforms the previous best model by
            over 20\% across all evaluation metrics on CN-82K.
            For ConceptNet datasets, without using the textual information, ConvE
            and RotatE perform better than ComplEx and DistMult.
            \cite{malaviya2020commonsense} outperforms ConvE by a large margin with
            BERT features as initialization. This indicates that semantic
            information plays an important role for ConceptNet entities.
            
            For ATOMIC, ComplEx and
            InductivE provide the best performance among all methods. Text attribute feature is less effective than that for the ConceptNet
            dataset. We observe that the relation types from ATOMIC
            (e.g.  xAttr, xIntent, oReact) are more complex and require more high-level
            reasoning. Thus, it is more difficult to infer directly from semantic
            embeddings. As compared with ComplEx, InductivE is good at HITs@10
            score, which means more entities are ranked higher. This is attributed
            to synthetic edges that help entities with limited connections to get
            reasonable performance as more synthetic connections are added.

        \subsubsection{Inductive link prediction}
        
            Table \ref{tab:inductive} summaries results on CN-82K-Ind and ATOMIC-Ind. From this table, we observe that: 
            
            \begin{itemize}
            \item The conventional KG embedding models, ConvE and RotatE, perform badly in our proposed inductive settings.
            ConvE and RotatE learn entity embeddings via learning to score positive/negative links between entities. After training, unseen entities remain to be randomly initialized, because no links over unseen entities can be observed during training. These random embeddings for unseen entities explain the poor performance of ConvE and RotatE.
            \item Comparing with
            \cite{malaviya2020commonsense}, InductivE achieves the best performance on both inductive datasets. It has an improvement of 48\% on MRR when comparing with \cite{malaviya2020commonsense} on CN-82K-Ind dataset. As shown in the transductive result (Table \ref{tab:transductive}), text features for ATOMIC dataset are less effective than that for ConceptNet dataset, thus results in much worse performance on ATOMIC-Ind than on CN-82K-Ind.
            \end{itemize}
            
            
           In \cite{malaviya2020commonsense} the model training adopts a two-branch training process:
           in the GCN branch, node features are randomly initialized; in the text encoder branch, entities embeddings are \emph{initialized} by BERT features then finetuned during training.
           For both branches, training \emph{does not} change the embeddings for unseen entities, because no links over unseen entities can be observed during training. Therefore, at test time, embeddings for unseen entities are computed from random node features (in the GCN branch) plus raw BERT features (in the text encoder branch), which are different from those of trained seen entities, thus results in poor performance.

           In contrast, InductivE first ensures inductive learning ability by using fixed text embedding as the input to GR-GCN model and further enhances the neighboring structure for unseen entities via the proposed graph densifier. Both modules contribute to the good performance of InductivE.

            To conclude,
            InductivE is the first benchmark in the inductive setting while
            providing competitive performance in the transductive setting. 

        \begin{table}[htb]
        \centering
        \caption{Ablation study on CN-82K-Ind dataset.}
        \resizebox{0.45\textwidth}{!}{
        \begin{tabular}{ l | c  c     }\hline
                           \multirow{2}{*}{\textbf{Model}}      & \multicolumn{2}{c }{\textbf{CN-82K-Ind}}                \\
                                                            & MRR          & Hits@10       \\ \hline
         \textbf{InductivE}     & 18.15    & 29.37     \\
         replace GR-GCN with MLP & -1.70   & -2.60 \\
         remove graph densifier &  -3.03  & -4.60 \\
         remove gating in GR-GCN & -3.79 & -5.37 \\\hline
         BERT feature only & 16.57  & 27.68\\
         fastText feature only & 16.14 & 26.87\\
          \hline
        \end{tabular}
        }
        \label{tab:ablation}
        \end{table}

        \subsection{Ablation Study and Analysis}
        \label{sec:abla}
                
            \subsubsection{Ablation study}
            
            To better understand the contribution of different modules in InductivE to the performance, we present ablation studies in Table \ref{tab:ablation}.
            
            
            \begin{itemize}
                \item \textbf{Feature Analysis}: The performance gap between BERT feature and fastText     feature is not huge. It is because the text attributes for CKG datasets are short phrases, with an average of 2.9 words for ConceptNet and 4.4 words for ATOMIC. BERT is more powerful for sentence-level sequences and that is why we use a concatenation of features to have multi-view entity representations.
                \item \textbf{Replace GR-GCN with MLP}: The MRR score drops from 18.15 to 16.45. This indicates that learning from the graph structural information can further boost the performance, although our free-text encoder provides a strong baseline and most essential for inductive learning.
                \item \textbf{Remove graph densifier}: This means the
                model only learns from the original triplet graph structure. The MRR
                score drops from 18.15 to 15.12. This indicates our graph densifier is helpful for inductive learning, since that the added synthetic edges are particularly helpful for
                unseen entities since they do not have existing neighboring structure to learn
                from at the beginning.
                \item \textbf{Remove gating in GR-GCN}: The performance drops from 18.15 to 14.36. This demonstrates the importance of
                the gating function, which adaptively controls the amount of neighboring
                information flowing into the center node. Neighboring nodes in CKGs are diverse and connected to the center node with different
                relations. Without the gating function, different information sources
                are injected directly to the center node which can cause confusion to the
                model.
            \end{itemize}

        \subsubsection{Analysis on graph densifier} Our iterative graph densifier is a special design to provide more structural information for unseen entities which does not contain any existing links at the beginning. In the meantime, there are other alternatives \cite{malaviya2020commonsense} to build syntactic links such as directly building syntactic links from raw features. Even though constructing synthetic links directly from raw features is easier to implement, it has two disadvantages when targeting on our inductive learning framework: 1) A low threshold leads to noisy synthetic links and a high threshold provide little extra information as the feature has served as the input. 2) Synthetic links can be unbalanced between entities and some unseen entities will still have no connection even after densification.
        
        To verify our graph densifier is superior than its alternatives, we compare it with two alternatives:
        \begin{itemize}
            \item Raw feature with global thresholding (GS)\cite{malaviya2020commonsense}: The BERT feature is taken and compute pairwise cosine similarity between entities. A global threshold (0.95) is set for the whole graph and similarity links are added between entities if their cosine similarity is above the threshold.
            \item Raw feature with fixed neighboring (FN): To make sure that all entities get a reasonable amount of neighboring information, we add similarity links to entities until they have at least 5 neighboring entities. The candidates are selected by ranking cosine similarity of the BERT feature.
        \end{itemize}
        
        \begin{table}[htb]
        \centering
        \caption{Analysis on our graph densifier}
        \resizebox{0.40\textwidth}{!}{
        \begin{tabular}{ l | c  c   }\hline
        \multirow{2}{*}{\textbf{Model}}      & \multicolumn{2}{c }{\textbf{CN-82K-Ind}}                \\
        & MRR          & Hits@10       \\ \hline
         Our graph densifier & \textbf{18.15} &  \textbf{29.37}     \\
         Raw feature with GS & 13.25  & 22.98 \\
         Raw feature with FN & 10.12  & 17.04 \\\hline
        \end{tabular}
        }
        \label{appendix:densifier}
        \end{table}
        
        The results are shown in Table \ref{appendix:densifier}. Our graph densifier achieves the best performance. In comparison, \textit{Raw feature with GS} is not providing satisfactory result because the raw feature can be noisy and imprecise links can be added. Even though some high-quality similarity links are constructed, it does not provide much extra information. \textit{Raw feature with FN} is the worst choice because adding fixed number of links directly from raw feature will allow more noisy links injected to the graph structure which leads more confusion in general. Since our graph densifier incorporates an iterative approach, the features we use are generally more fine-tuned with current datasets and keep improving during training and thus providing synthetic links with a higher quality.

        \begin{table}[htb]
        \centering
        \caption{Top-3 nearest neighbors of unseen entities 
        }
        \vspace{1 mm}
        \resizebox{0.48\textwidth}{!}{
        \begin{tabular}{ l || l | l }
        \hline 
        \textbf{Unseen entity} & \textbf{Ours}  & \textbf{Malaviya et al.} \\
        \hline 
        
        \multirow{3}{*}{pay for subway}         & pay subway fare               & pay for drink     \\
                                                & pay traffic ticket            & pay on billet     \\
                                                & pay fare                      & pay for package   \\ \hline

        \multirow{3}{*}{perform experiment}     & try experiment                & do experiment\\
                                                & do experiment                 & conduct experiment\\ 
                                                & \textbf{test his hypothesis}    & run experiment \\ \hline
        
        \multirow{3}{*}{item fill with air}     & inflate thing with air             & open container    \\ 
                                                & \textbf{balloon fly because they}   & wled like item   \\ 
                                                & \textbf{balloon go up because they}  & microscopic thing \\ \hline

        \multirow{3}{*}{wait for you airplane}                 & get prepare to wait             & wait for while    \\ 
                                                & \textbf{run short of fly}   & wait for windy day   \\ 
                                                & \textbf{transport passenger} & wait for blue bird \\ \hline 

        
        \end{tabular}
        }
        \label{tab:visul}
        \end{table}
        \subsubsection{Case study on graph densifier}

            By examining our graph densifier carefully, we notice some interesting phenomena associated with
            synthetic similarity links. In Table \ref{tab:visul}, we list top-3 nearest neighbors of unseen entities. As expected, some similarity relations can be discovered with our graph densifier. For example,  \emph{``pay for subway''} shares almost the same semantic
            meaning with \emph{``pay subway fare''}.
            Therefore, when computing embedding for \emph{``pay for subway''}, the
            embedding for \emph{``pay subway fare''}
            can be a reliable reference.  Unexpectedly, other complex relation
            types can also be discovered using our densifier. Entities marked as bold in Table \ref{tab:visul} indicates the candidate
            entity has more complex relation with the unseen entity. 
            For example, \emph{``perform experiment''} is the \emph{``Goal''} to \emph{``test his hypothesis''}. The
            \emph{``Reason''} for \emph{``wait for you airplane''} is
            \emph{``run short of fly''}. 
            With rich
            local connectivity, unseen entities can perform multi-hop reasoning over
            graphs and obtain high quality embeddings. Comparing with previous method \cite{malaviya2020commonsense}, our proposed graph densifier can construct higher quality synthetic edges. This could be especially helpful for unseen entities whose neighboring structures are built using our graph densifier.

\section{Related Work}
    
\subsubsection{Knowledge Graph Embedding}
Knowledge graph completion by predicting missing links has been
intensively investigated in recent years. Most methods are
embedding-based.
TransE \cite{bordes2013translating} models the relationship
between entities with nice translation property ($e_h + e_r \approx
e_t$) in the embedding space. ComplEx \cite{trouillon2016complex} and
RotatE \cite{sun2019rotate} represent embeddings in a complex space to model more complicated relation interactions. Instead of using a
simple score function, ConvE \cite{convE} applied convolution to
embedding so as to allow more interactions among triplet features. To
exploit the structural information, GNNs are applied to multi-relational
graphs as done in R-GCN \cite{schlichtkrull2018modeling} and SCAN
\cite{shang2019end}. All above-mentioned methods learn embedding based
on a fixed set of entities and are transductive as originally proposed. 
    
\subsubsection{Inductive Learning on Graphs} 
Inductive learning is investigated in the last several years for both
graphs and knowledge graphs. GraphSage \cite{hamilton2017inductive}
relies on node features and learns a local aggregation function that is
generalizable to newly observed subgraphs. Most KGs embedding models are focusing non-attributed graph so that inductive learning is mostly relying on local
connections for unseen entities \cite{albooyeh2020out}. \cite{hamaguchi2017knowledge} generates
embedding for unseen nodes by aggregating the information from
surrounding known nodes. \cite{teru2019inductive} models relation
prediction as a subgraph reasoning problem. However, because unseen
entities in CKGs have no existing link, structural-based methods cannot be
applied \cite{teru2019inductive,albooyeh2020out}. %
As an alternative, there exists work that incorporates
entity description in the embedding process and can be inductive
by nature. For example, \cite{xie2016representation} learns a joint
embedding space for conventional entity embedding and description-based
embedding. 
Our work exploits both
structure information and textual description for inductive purpose on CKGs. 
    

\subsubsection{Language model on CKGs} 
Recently, researchers attempt to link commonsense knowledge with
pre-trained language models \cite{petroni2019language,yao2019kg,devlin-etal-2019-bert}. 
COMET \cite{Bosselut2019COMETCT} is a generative model that transfers
knowledge from pre-trained language models and generates new facts in
CKGs.  It could achieve performance close to human beings. However, COMET always introduces
novel/unseen entities to an existing graph, leading to an even sparser
graph.  With unseen entities constantly
introduced by generative models, an inductive learning method, such as
InductivE, is important in practice. 
            
\section{Conclusion}

    In this work, we propose to study the inductive learning problem on CKG completion, where unseen entities are involved in link prediction. To better evaluate CKG completion task in both transductive and inductive settings, we release one new ConceptNet dataset and two inductive data splits for future research and development purposes. Dedicate to this task, a new embedding-based framework InductivE is proposed as the first benchmarking on inductive CKG completion. InductivE leverages entity attributes with transfer learning and considers structural information with GNNs. Experiments on both transductive and inductive settings show that InductivE outperformed the state-of-the-art method considerably.

    Inductive learning on CKG completion is still at its infancy. There are many promising directions for future work. For example, large pre-trained language models (LMs) have shown effective in capturing implicit commonsense knowledge from large corpus. 
    How to effectively merge the knowledge in large pre-trained LMs with structured CKG could be an interesting direction to explore.
    Another direction is to explore inductive learning on unseen/new relations that are truly useful in real CKG expansion task.

\bibliographystyle{IEEEtran}
\bibliography{custom}

\end{document}